# From Bytes to Bites:
# Using Country Specific Machine Learning Models
# to Predict Famine

Salloni Kapoor[1], Simeon Sayer[2]
[1] Issaquah High School, Issaquah, WA, 98027, United States
[2] Faculty of Arts and Sciences, Harvard University, Cambridge, MA, 02138, United States
kapoorsalloni5@gmail.com; ssayer@fas.harvard.edu

*Hunger crises are critical global issues affecting millions, particularly in low-income and developing countries. This research investigates how machine learning can be utilized to predict and inform decisions regarding famine and hunger crises. By leveraging a diverse set of variables—natural, economic, and conflict-related—three machine learning models (Linear Regression, XGBoost, and RandomForestRegressor) were employed to predict food consumption scores, a key indicator of household nutrition. The RandomForestRegressor emerged as the most accurate model, with an average prediction error of 10.6%, though accuracy varied significantly across countries, ranging from 2% to over 30%. Notably, economic indicators were consistently the most significant predictors of average household nutrition, while no single feature dominated across all regions, underscoring the necessity for comprehensive data collection and tailored, country-specific models. These findings highlight the potential of machine learning, particularly Random Forests, to enhance famine prediction, suggesting that continued research and improved data gathering are essential for more effective global hunger forecasting.*

I.  **Introduction**

Famine has plagued human societies throughout history, and despite advances in technology and data analysis, predicting and preventing hunger crises remains a significant global challenge. Throughout the ages, people have turned to traditions and cultural references, such as Groundhog Day in the U.S. and Canada, to make sense of unpredictable conditions and forecast future food supply. More theoretical approaches have also been used as justification, such as the theories proposed by Thomas Malthus [1], suggesting that famine was inevitable as population growth would eventually surpass Earth's food supply. These ideas provided early frameworks for understanding food scarcity, acting as stepping stones for modern systems like FEWSNET (Famine Early Warning Systems Network) [2] which offer real-time data on food security. Despite their analytical approach, such systems still struggle with accurately predicting future conditions.

In recent years, machine learning has emerged as a promising tool for predicting and informing decisions regarding hunger crises. Machine learning models, particularly those using supervised learning and regression techniques, have the potential to enhance famine predictions by analyzing vast amounts of data and identifying patterns that might be missed by traditional methods. However, many of these approaches fail to fully integrate the available data for the region they are analyzing, creating weaker models. Additionally, the results of these models give little guidance into region specific priorities for improving future data collection, which is critical for refining and enhancing predictive accuracy.

This project builds on existing approaches for predicting future food insecurity by employing a regression model that leverages numerical data, including satellite imagery, food specific economic factors, and conflict-related statistics. Unlike previous papers that often focus solely on a specific region, or utilize a limited set of input variables, this approach integrates a broad array of data sources, but methodically selects the richest subset of that data to train a region specific model. This research proves that there is no universal formula for famine, and that important variables in one region are useless in another. By pairing these optimal regional inputs with the corresponding food consumption score [3]—a key indicator of food security ranging from 0 (worst) to 112 (best)—the model aims to produce more accurate and context-sensitive predictions, while simultaneously providing insight into the optimal famine indicators in the region. This helps to identify and address the complex and varied factors contributing to food insecurity across different countries.

These quantitative insights can guide humanitarian efforts and policy decisions. Using numerical data in a regression model enables precise and actionable predictions, offering a more reliable basis for decision-making compared to traditional forecasting methods.

Improving the accuracy and timeliness of famine predictions is crucial for effective intervention and resource allocation. Leveraging machine learning enhances current prediction models while also informing and reshaping future data collection strategies. Integrating diverse data sources, including economic indicators, environmental factors, and conflict statistics, into a single predictive model provides a comprehensive understanding of the multifaceted nature of food insecurity.

Accurately forecasting food consumption scores allows for more proactive and informed responses to emerging hunger crises. The insights gained could lead to more targeted and efficient humanitarian interventions, potentially mitigating the impact of famine on vulnerable populations worldwide.

## II. Background

Initial approaches to machine learning for famine detection involved using models to forecast food security transitions. The paper "Forecasting Transitions in the State of Food Security with Machine Learning Using Transferable Features" [4] showcases a method utilizing Extreme Gradient Boosting (XGBoost) to predict monthly changes in food security in Ethiopia. By integrating various open-source datasets, this model outperforms traditional methods by providing more frequent and accurate predictions, which are essential for timely humanitarian actions. The model's ability to generalize across different regions makes it a scalable and transferable solution for global food security forecasting. However, the model's effectiveness is highly dependent on the availability and quality of input data, and it may require significant computational resources.

A key methodology of addressing food insecurity using machine learning models was to analyze more localized or specific subsets of food security data was proposed in "Machine learning can guide food security efforts when primary data are not available" [5]. This method integrates diverse data sources like UN household surveys, remote sensing, and market analytics and applies a variety of model approaches, such as neural networks or support vector machines.

The key contribution of this approach is its ability to provide detailed insights into specific communities or regions, enabling more tailored predictions and early warning systems. The accuracy of the predictions was also impressively high. While this method enhances predictive accuracy by incorporating various data sources, these data sources may not exist in all regions, and so it may struggle with scalability across different countries and in places calls for highly specialized data that may not be universally available, or applicable.

Both approaches play a significant role in the field of hunger crisis prediction using machine learning. The first approach emphasizes the scalability and transferability of models like XGBoost, making them suitable for broader regional predictions. In contrast, the second approach offers specialized insights into specific aspects of food security, which can be particularly valuable in localized or resource-constrained settings. The XGBoost model follows the principles of gradient boosting, optimizing predictive power by combining multiple weak learners into a strong ensemble. SHAP (SHapley Additive exPlanations) values are used to interpret the model's predictions, providing insights into the importance of each feature. The results show that the XGBoost model [5] explained up to 81% of the variation in insufficient food consumption and 73% of the variation in food-based coping levels, demonstrating its strong predictive capabilities. This study highlights the critical importance of data collection and the need to expand the range of data sources to improve the accuracy of food security predictions. In summary, while the first approach offers a scalable solution for regional forecasting, the second provides granular insights crucial for addressing specific local challenges. Both methods underscore the growing importance of machine learning in guiding global food security efforts, and further research and data collection are necessary to refine these models and enhance their applicability across diverse contexts.

### III. Dataset

The dataset utilized is a comprehensive compilation of numerical data from a variety of verified sources, encompassing environmental, economic, and conflict-related variables. These sources include international organizations, independent NGOs, and scientific research bodies, providing a reliable and multifaceted foundation for the analysis. As described in Table 1, the data includes critical metrics such as rainfall patterns, inflation rates, and conflict statistics over various time periods, all of which are essential for predicting food security outcomes.

One of the primary sources of data is the Armed Conflict Location and Event Data Project (ACLED) [6], which provides detailed information on conflict-related events. This includes the number of fatalities from battles, violence against civilians, and remote violence, all reported over a rolling 90-day period with a 14-day lag. This conflict data is crucial for understanding the direct impact of violence on food security, as regions experiencing high levels of conflict often face disruptions in food production and distribution.

Another significant source is FAOSTAT, the statistical division of the Food and Agriculture Organization of the United Nations [7]. FAOSTAT provides data on the Prevalence of Undernourishment (PEWI), calculated as the mean, minimum, and maximum values over the last three months. These metrics offer vital insights into the nutritional status of populations and help assess the severity of food insecurity in different regions.

Economic data is sourced from Trading Economics [8], which offers key indicators such as headline inflation rates, food inflation rates, and currency exchange rate variations. These economic variables are essential for understanding the pressures that influence food availability and affordability, particularly in regions where high inflation can significantly erode purchasing power and exacerbate food insecurity.

Environmental data is drawn from the World Food Programme's (WFP) Seasonal Monitor [9], which uses data from the Climate Hazards Group InfraRed Precipitation with Station data (CHIRPS) [10] and the Moderate Resolution Imaging Spectroradiometer (MODIS) [11]. This environmental data includes rainfall anomalies and Normalized Difference Vegetation Index (NDVI) values, which are critical indicators of agricultural productivity. For example, rainfall anomalies can signal potential droughts, while NDVI values reflect the health of vegetation and, by extension, the likely yield of crops.

*Table 1.* *General Feature Descriptions*

| Variable Type | Description | Source |
|---|---|---|
| **Population Density** | Refers to the number of people living per unit of area, usually per square kilometer. Calculated by averaging all pixel values within each first-level administrative unit. | Center for International Earth Science Information Network (CIESIN) |

| **Rainfall Anomaly** | The deviation of observed rainfall from the long-term average. Positive anomalies indicate more rainfall than average, while negative anomalies indicate less. | WFP Seasonal Explorer, using CHIRPS data |
|---|---|---|
| **Rainfall Value** | The actual measured amount of rainfall over a specific period. | WFP Seasonal Explorer, using CHIRPS data |
| **NDVI (Normalized Difference Vegetation Index)** | A measure of the health and greenness of vegetation, derived from satellite imagery. Higher NDVI values indicate healthier vegetation. | WFP Seasonal Explorer, using CHIRPS data |
| **NDVI Anomaly** | The deviation of NDVI values from the historical average, indicating changes in vegetation health compared to typical conditions. | WFP Seasonal Explorer, using CHIRPS data |
| **Prevalence of Undernourishment (PEWI)** | A measure representing the percentage of a population whose food intake is insufficient to meet dietary energy requirements. | FAOSTAT |
| **Headline Inflation** | The overall inflation rate, capturing the increase in the general price level of goods and services over time. | Trading Economics |
| **Food Inflation** | The rate at which the prices of food items are increasing over time. | Trading Economics |
| **Currency Exchange Rate Variation** | The percentage change in the currency exchange rate over a specific period, reflecting how the value of a currency has changed relative to others. | Trading Economics |
| **Battle Fatalities** | The number of deaths resulting from direct combat between armed groups. | Armed Conflict Location and Event Data Project (ACLED) |
| **Remote Violence Fatalities** | The number of deaths caused by violence executed from a distance, such as airstrikes, artillery, or drone attacks. | Armed Conflict Location and Event Data Project (ACLED) |
| **Violence Against Civilians Fatalities** | The number of deaths resulting from deliberate violence against non-combatant civilians. | Armed Conflict Location and Event Data Project (ACLED) |
| **Overall Conflict Fatalities** | The total number of fatalities from all types of conflict events, including battles, remote violence, and violence against civilians. | Armed Conflict Location and Event Data Project (ACLED) |

Given the diverse origins of the data, significant preprocessing was necessary to ensure consistency and usability across all features. This preprocessing involved managing missing data, particularly in instances where certain variables were not available for all countries. For example, in regions with low levels of violence, the ACLED may not have comprehensive data on conflict fatalities, necessitating the use of imputation techniques or alternative data sources to address these gaps.

The significance of each feature within the dataset is evident in its contribution to predicting food security outcomes. Conflict data from ACLED captures the impact of violence on food security, as areas with high conflict often experience severe disruptions in food supply chains. Economic indicators from Trading Economics reflect the broader economic environment, where variables like inflation directly affect the affordability of food. Environmental data from WFP's Seasonal Explorer provides crucial insights into climatic conditions that influence crop yields, with anomalies in rainfall and NDVI serving as early warnings for potential food shortages.

## IV. Methodology

The initial steps needed for the development of a model was the consolidation of the data. The final data was represented by a mixture of countries, each with its own set of possible features. Additionally, the time resolution of this data varied - some looked at the past 3 months, others 90 days, others still as an average of the previous year. Overall, the model had access to 31 variables, over a total of over 73,000 entries, across 69 countries. However, very few countries had good data over sufficient time periods for all of these variables. To require the model to train on all variables would mean discarding large amounts of valuable data from countries that were, for instance, missing only a single variable. This meant that country specific models were significantly better suited to the task, as it allowed a model to independently evaluate a country based on the data available, and not be influenced by what data was available in other countries. This solution was not perfect in ensuring data consistency - for instance, in a given country, for a given year, the data collected was not always the same. It is possible that for three of the ten years, there was substantial data collection of battle fatalities due to a conflict in the region, but this data did not exist for the other years. The dataset was tailored for each country by including only the features available for the majority of that country's data when training the model. This prevents a single feature with a small representation 'hijacking' the dataset.

Once this data was isolated, each country was split into a testing and training group, with an 80:20 train-test split, with the survey variables being the input, and the FCS score being the output. The training data was used to fit the various models we experimented with, and the testing data was used to generate validation accuracies of those models. The mean absolute error between the predictions of the model on the test data and the real values was computed to evaluate the error of a given model for a given country.

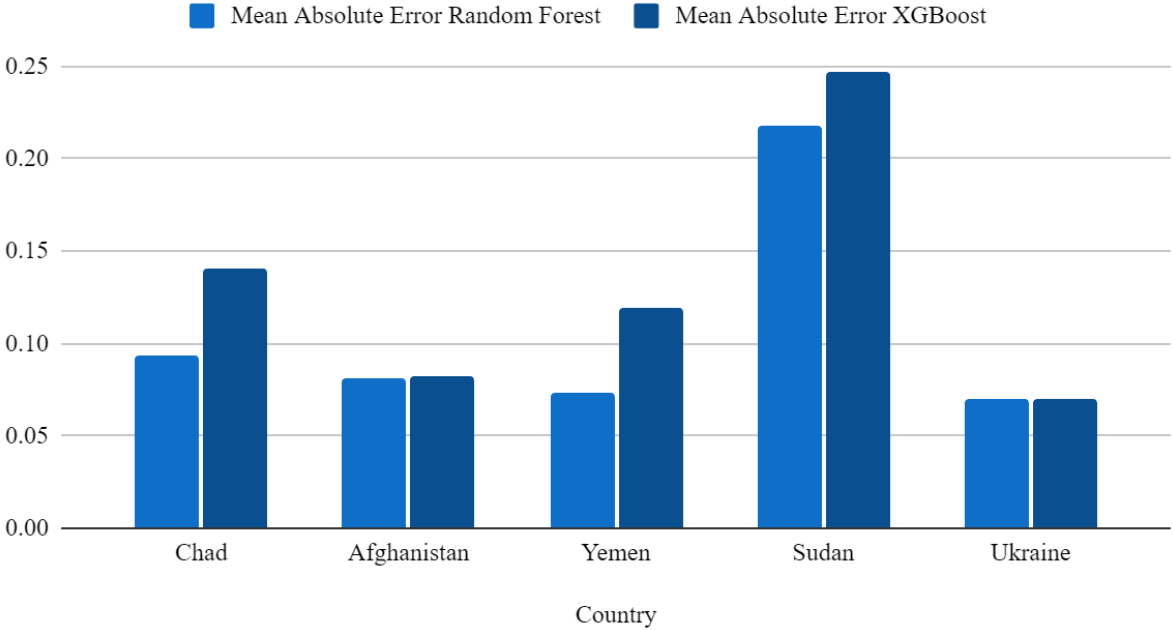

*Fig 1.* *Random Forest Error vs. XGBoost Error*

We systematically tried a variety of models, with particular focus on XGBoost, an advanced gradient boosting algorithm known for its high performance and efficiency, as highlighted in existing literature. However, as seen in Figure 1, this research found that Random Forest, an ensemble learning method that combines multiple decision trees, performed as well or better than XGBoost. Additionally, Random Forest provides more easily interpretable model weights and feature importance, making it a clearer approach compared to calculating SHAP values. This clarity made it more efficient for evaluating model priorities and provided results that were more easily understood by policymakers and the public. Other models, such as linear regression, performed poorly, and some, like neural networks, were both poor predictors and difficult to interpret, so they were discarded.

Using Random Forest, and the initial findings of the feature importances from that regression model, a categorization could be made regarding which variables were most important to determining that country's FCS score. The rankings of the features were used to create a scoring system that averaged different categories of variables, and saw which category of variables were most important overall to the model's decision. Each variable from our input was assigned a group, based on the fundamental aspect of the country it measured. For instance, "rainfall" was given a natural label, and "battle fatalities" were given a combat label. If, on average, the variables pertaining to economic factors were most useful, and therefore highest ranked, the country would be designated an economic famine category. It was therefore possible to categorize the nature of each region's famine - in our case, either natural, economic or combat factors. Although most countries had a mix of important variables, being able to come up with relevant groupings and designations, and have these groupings be quantified is important to interpreting the results of this model, future research in the area, and enacting policy action. The breakdown of the proportion and location of these are shown below in Figure 2 and Figure 3.

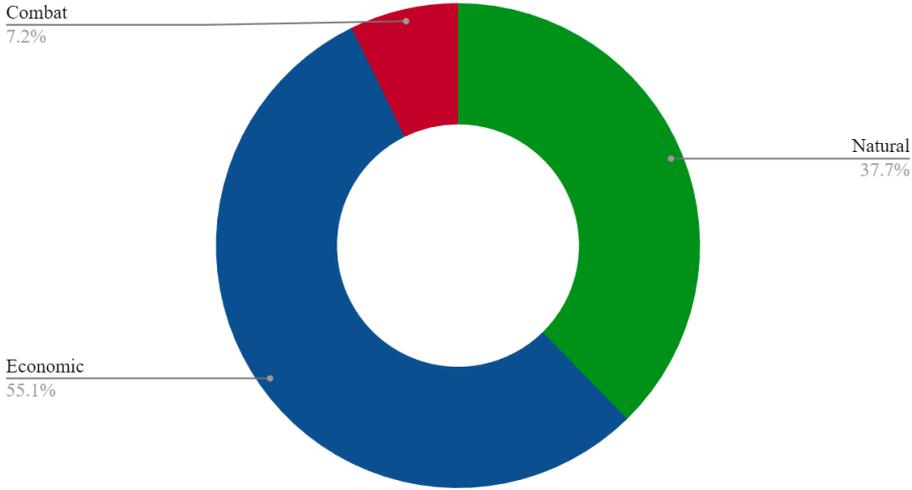

*Fig 2.* *Proportion of Country Famine Categories, according to model*

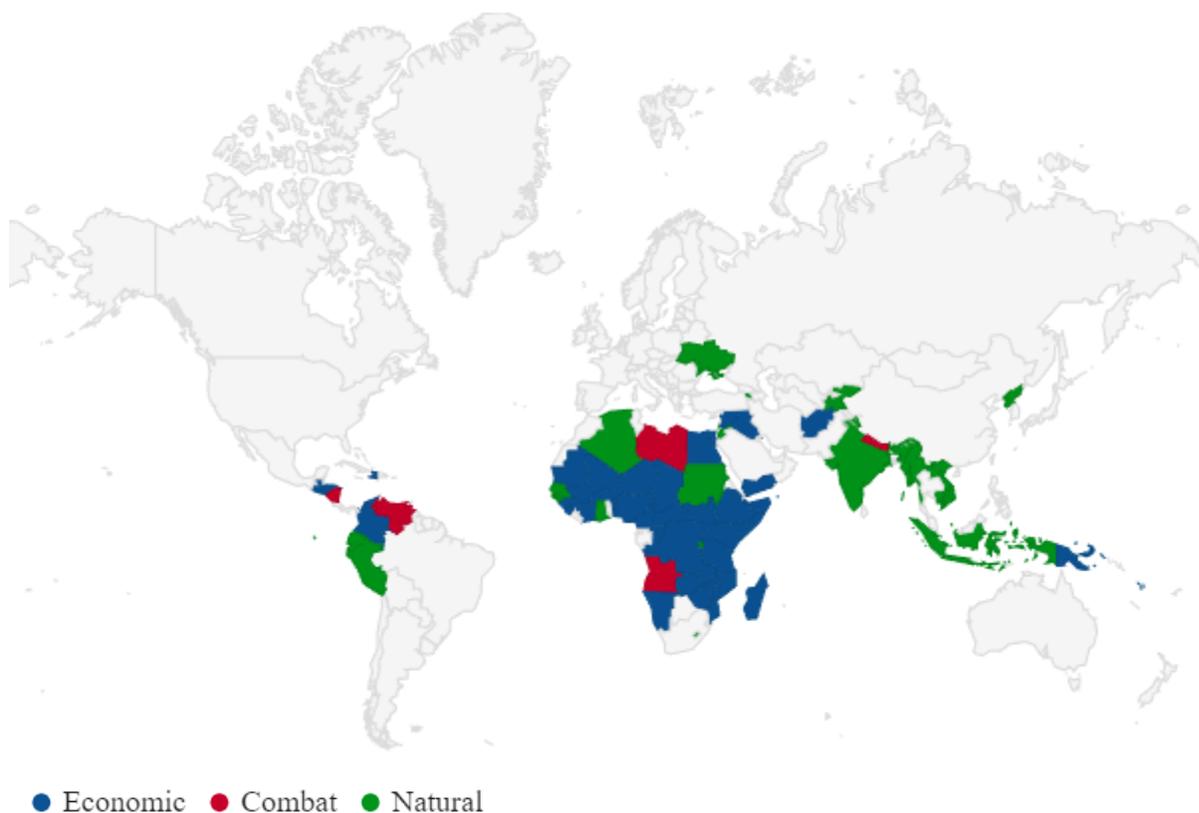

***Fig 3.*** *Location of Country Famine Categories, according to model*

## V.  Results and Discussion

The chosen Random Forest model performed strongly, with an average absolute error for FCS score of 10.6%. In the context of the predicted variables, this means that the model is able to estimate the percentage of people with poor or borderline food consumption in an area with an average deviation of 10.6 percentage points from the actual values. Some countries saw significantly better results—Indonesia, for example, had an error rate as low as 2%. However, South Sudan struggled, with a much higher error rate of 30%, indicating considerable fluctuations or factors beyond what was captured by our data.

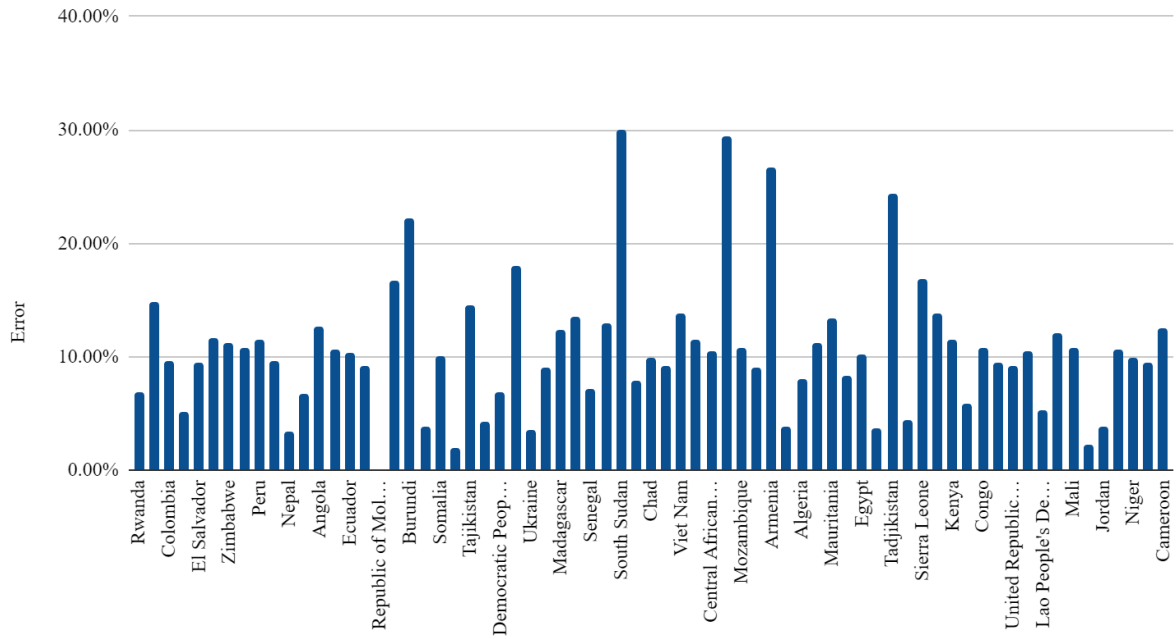

***Fig 4***. *Mean Absolute Error by Country*

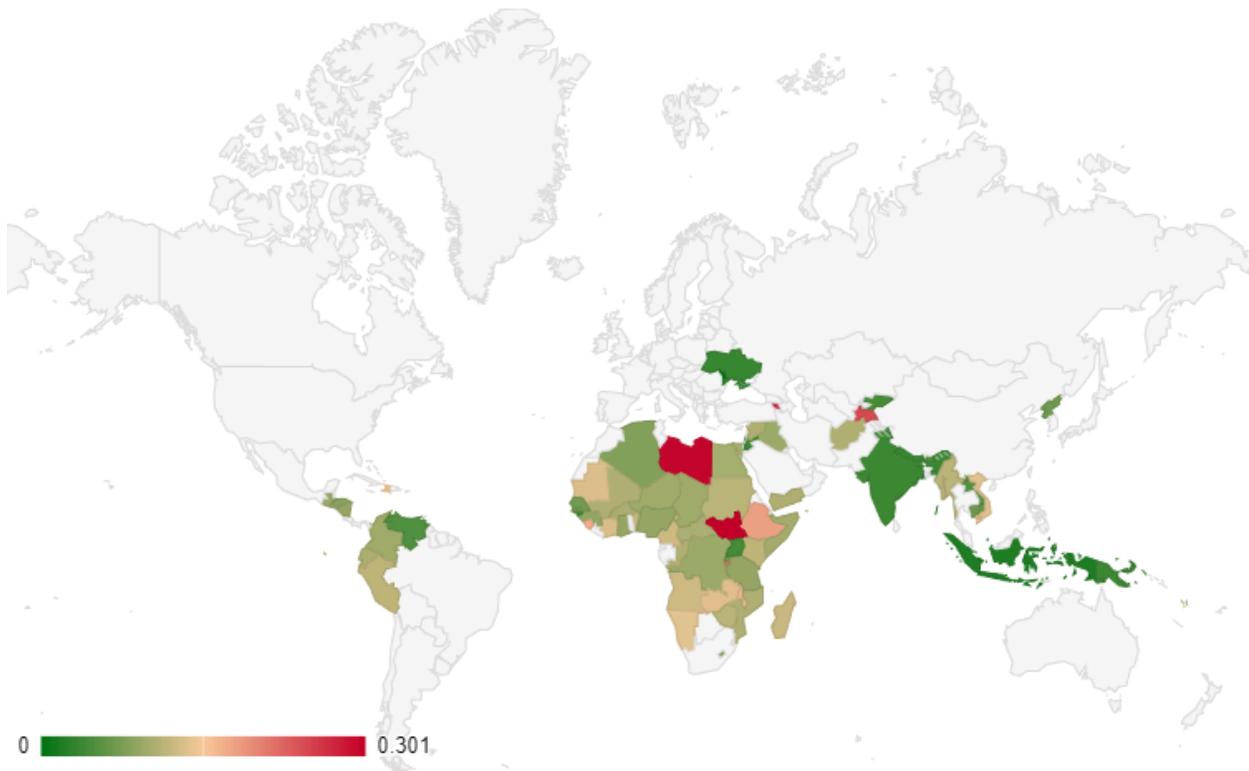

***Fig 5.*** *Global Error Visualized*

Analyzing these results reveals interesting trends linking a country's model accuracy to the relative importance of specific features used in the predictions. The Random Forest model, in particular, allowed for identifying which features were most influential. In many of the top-performing countries, models heavily relied on economic indicators like inflation rates. Conversely, countries with poorer performance, such as South Sudan, had models that depended more on conflict-related data, such as recent battle fatalities. These are general trends, with notable exceptions—Nepal, for instance, achieved an impressively low average error of 3.44%, despite its model primarily using battle fatality data.

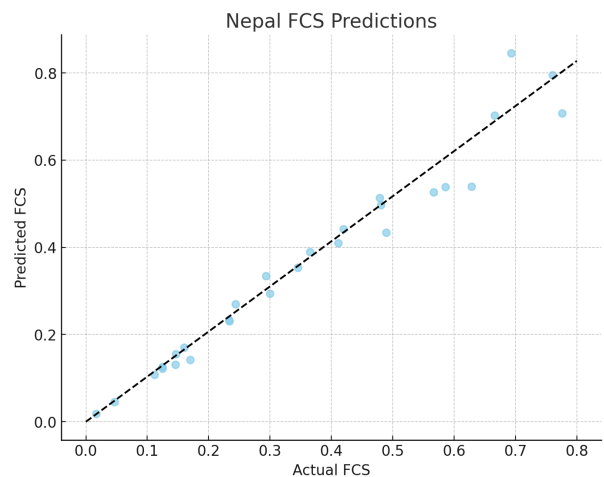

*Fig 6.* Actual vs. Predicted Graph for Nepal Food Consumption Scores. (Line y = x shown in black, for reference)

This prompted an enhanced analysis on which factors were most useful in making predictions relating to famine. As seen in Figure 7, when analyzed, every single feature that was available to the model at some point ranked the top five *most important* for at least one country. Every feature also appeared in the *bottom* four most important features for at least one country. There may be explanations for these results - it is possible that with many features being targeted at measuring similar areas, that feature redundancy caused low ranking of features when a proxy could be used instead. However, it is clear that there are no individual variables that are absolute winners for the random forest model's prediction abilities.

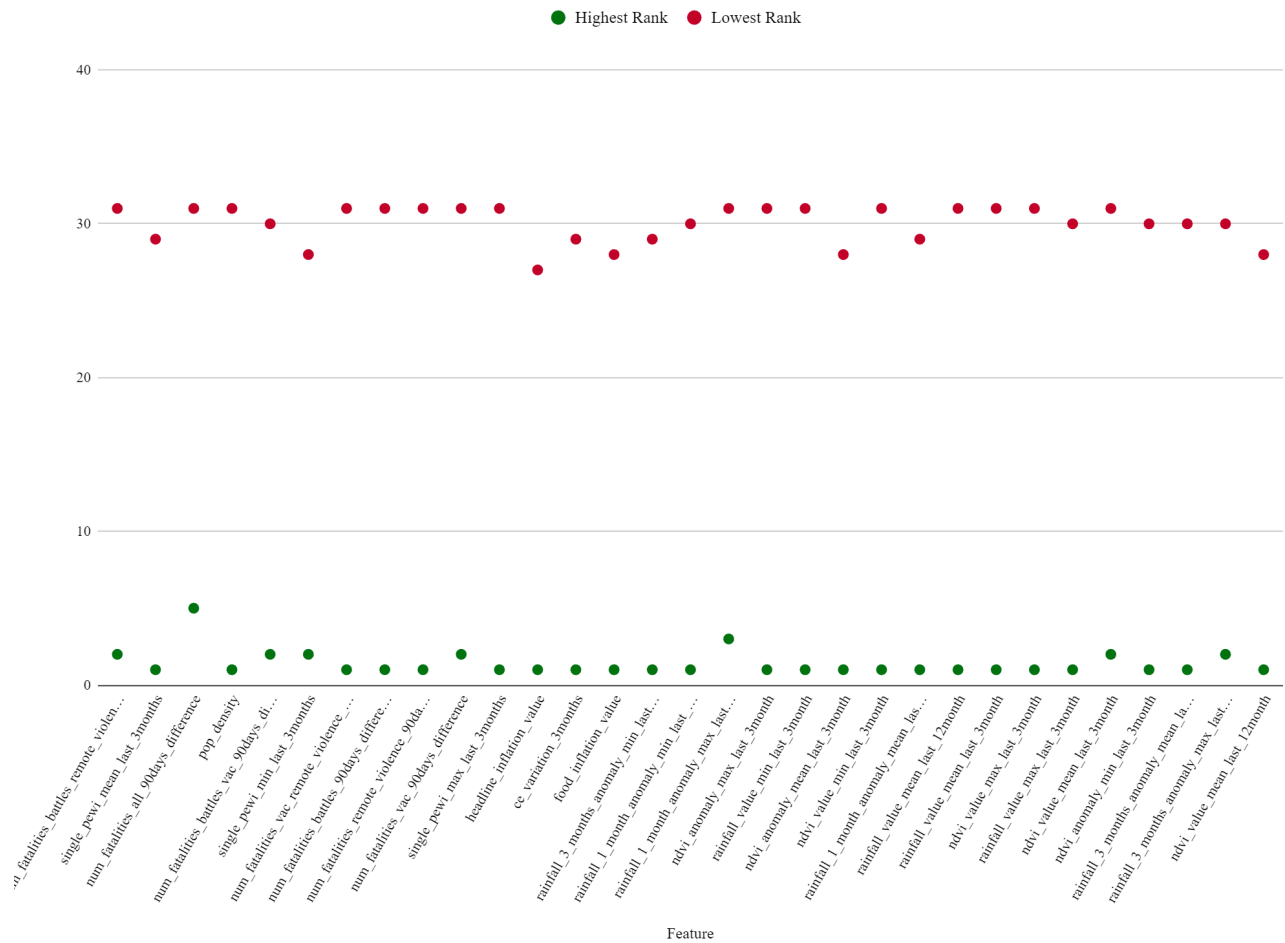

*Fig 7.* *Feature Importance Variation*

A sample of the feature importances for a select group of countries appears in Table 2. Observe the variety of features, used in a variety of orders, emphasizing the fact that across countries, there is no single most important feature to predict FCS.

*Table 2. Feature Variation Breakdown*

| Country | Feature Importances (from Random Forest Regressor) |
|---|---|
| **Sudan** | Feature importances bar chart. Top features: rainfall_value_min_last_3month (~0.115), num_fatalities_battles_90days_difference (~0.10), ndvi_value_max_last_3month (~0.08), rainfall_1_month_anomaly_mean_last_3month (~0.075), num_fatalities_battles_remote_violence_90days_difference (~0.07), rainfall_value_mean_last_3month (~0.065), ndvi_anomaly_min_last_3month (~0.05), ndvi_value_mean_last_3month (~0.045), rainfall_3_months_anomaly_min_last_3month (~0.04), num_fatalities_vac_90days_difference (~0.04), rainfall_value_max_last_3month (~0.035), num_fatalities_battles_vac_90days_difference (~0.03), rainfall_1_month_anomaly_max_last_3month (~0.025), single_pewi_min_last_3months, rainfall_3_months_anomaly_max_last_3month, prevalence_of_undernourishment, rainfall_3_months_anomaly_mean_last_3month, num_fatalities_all_90days_difference, single_pewi_mean_last_3months, num_fatalities_vac_remote_violence_90days_difference, ce_variation_3months, pop_density, ndvi_anomaly_max_last_3month, single_pewi_max_last_3months, ndvi_value_min_last_3month, rainfall_value_mean_last_12month, ndvi_value_mean_last_12month, last_rcsi, ndvi_anomaly_mean_last_3month, headline_inflation_value, food_inflation_value, rainfall_1_month_anomaly_min_last_3month, num_fatalities_remote_violence_90days_difference. X-axis: Relative Importance (0.00–0.12). |
| **Haiti** | Feature importances bar chart. Top features: headline_inflation_value (~0.30), single_pewi_max_last_3months (~0.13), single_pewi_mean_last_3months (~0.075), food_inflation_value (~0.065), ndvi_anomaly_mean_last_3month (~0.04), pop_density, ce_variation_3months, rainfall_1_month_anomaly_min_last_3month, num_fatalities_battles_90days_difference, rainfall_value_mean_last_12month, rainfall_3_months_anomaly_max_last_3month, rainfall_1_month_anomaly_max_last_3month, num_fatalities_battles_vac_90days_difference, ndvi_value_mean_last_12month, num_fatalities_vac_remote_violence_90days_difference, ndvi_anomaly_max_last_3month, rainfall_value_min_last_3month, rainfall_3_months_anomaly_mean_last_3month, num_fatalities_vac_90days_difference, ndvi_anomaly_min_last_3month, num_fatalities_all_90days_difference, num_fatalities_battles_remote_violence_90days_difference, single_pewi_min_last_3months, rainfall_1_month_anomaly_mean_last_3month, rainfall_value_max_last_3month, rainfall_3_months_anomaly_min_last_3month, ndvi_value_min_last_3month, ndvi_value_max_last_3month, ndvi_value_mean_last_3month, rainfall_value_mean_last_3month, num_fatalities_remote_violence_90days_difference. X-axis: Relative Importance (0.00–0.30). |
| **Ukraine** | Feature importances bar chart. Top features: num_fatalities_battles_90days_difference (~0.12), rainfall_3_months_anomaly_min_last_3month (~0.075), rainfall_1_month_anomaly_max_last_3month (~0.065), ndvi_value_max_last_3month (~0.06), rainfall_1_month_anomaly_min_last_3month (~0.06), rainfall_value_max_last_3month (~0.055), single_pewi_mean_last_3months (~0.055), rainfall_3_months_anomaly_max_last_3month (~0.045), ndvi_anomaly_max_last_3month (~0.04), num_fatalities_vac_remote_violence_90days_difference, ndvi_value_min_last_3month, ndvi_anomaly_min_last_3month, num_fatalities_vac_90days_difference, ndvi_value_mean_last_12month, num_fatalities_battles_vac_90days_difference, single_pewi_max_last_3months, ndvi_anomaly_mean_last_3month, ndvi_value_mean_last_3month, food_inflation_value, rainfall_3_months_anomaly_mean_last_3month, pop_density, rainfall_1_month_anomaly_mean_last_3month, num_fatalities_battles_remote_violence_90days_difference, rainfall_value_mean_last_3month, headline_inflation_value, num_fatalities_remote_violence_90days_difference, rainfall_value_mean_last_12month, ce_variation_3months, single_pewi_min_last_3months, rainfall_value_min_last_3month, num_fatalities_all_90days_difference. X-axis: Relative Importance (0.00–0.12). |

| Colombia | 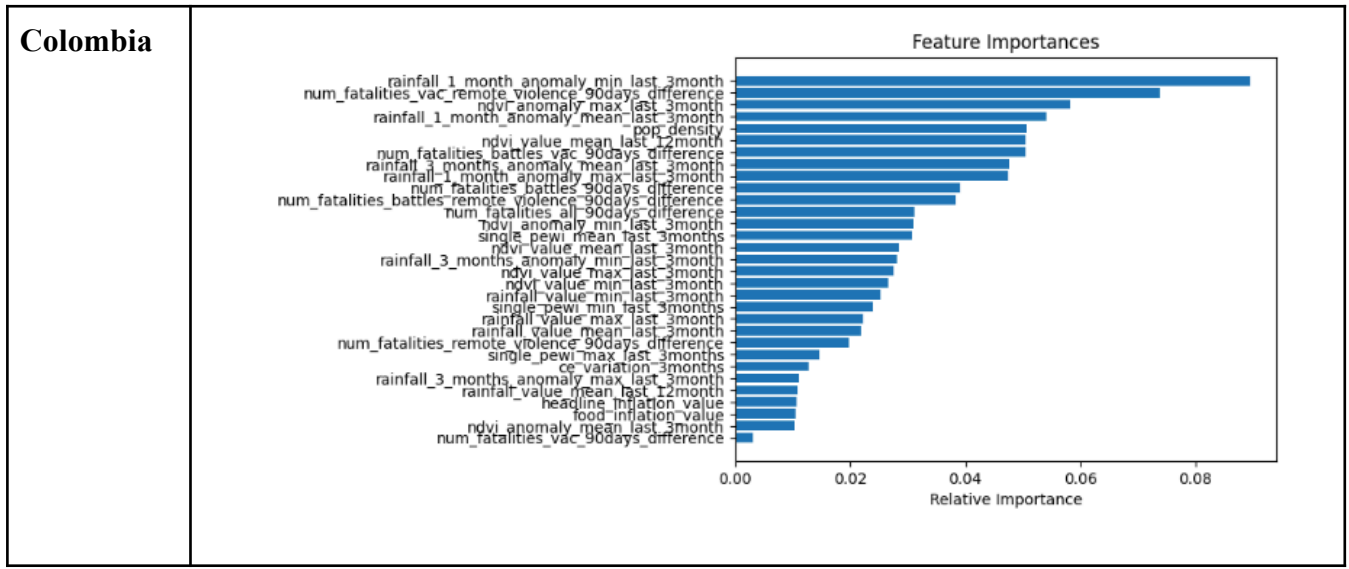 |
|---|---|

The results of this experiment are therefore three fold. First, this paper verifies that machine learning to project famine can be extremely reliable, in certain cases. The models best equipped to do this prediction are high depth decision trees, and similar models, like the random forest. The utilization of machine learning in providing humanitarian forecasting is both possible and necessary in the future of global aid. Second, this paper systematically proves the importance of data collection, the relevance of all existing data, and the necessity to expand both the collection and accessibility of data in this area. Third, machine learning researchers and global policy makers alike should be wary of drawing comparisons between humanitarian strategies. It is clear that, while there are common groupings, the data and therefore the root causes of famine vary wildly from country to country. This further promotes the need for widespread and detailed data from countries at risk of famine, in order to best forecast and ultimately address global famine, and for research to continue in the optimal way to process and analyze that data to make reliable predictions.

VI. **Conclusion**

Utilizing a supervised learning approach to predict a country's future food consumption score provides a quantitative measure that can guide humanitarian efforts and policy decisions. The regression model, based on numerical data, delivers precise and actionable predictions, offering a more reliable foundation for decision-making compared to traditional forecasting methods.

Improving the accuracy and timeliness of famine predictions is crucial for driving effective intervention and resource allocation. This machine learning approach not only sharpens current prediction models but also guides the evolution of future data collection strategies. By weaving together diverse data sources, such as economic indicators, environmental factors, and conflict statistics, into a cohesive predictive model, a deeper and more nuanced understanding of food insecurity emerges.

The power to accurately forecast food consumption scores unlocks the potential for more proactive and informed responses to unfolding hunger crises. The insights derived from this approach have the capacity to transform humanitarian interventions, making them more targeted and efficient, and ultimately reducing the devastating impact of famine on vulnerable communities worldwide.

## VII. Code and Data Availability

The raw data used to train this model is available as collected by the original paper [5]. The working code and supplementary table containing the full findings of all countries have been published alongside this paper at this link:
https://github.com/sallonikapoor/Bytes-to-Bites-Supplements.git

## VIII. Acknowledgements

Thank you to the many family members, friends, and colleagues who supported us in this research and helped us turn bytes into future bites.